\documentclass[runningheads]{llncs}
\usepackage{graphicx}
\usepackage{hyperref}

\usepackage[utf8]{inputenc}
\usepackage[misc]{ifsym} 
\usepackage{siunitx}
\begin{document}
\title{Biometric Fish Classification of Temperate Species Using Convolutional Neural Network with Squeeze-and-Excitation}
\titlerunning{Biometric Fish Classification of Temperate Species Using CNN-SENet}
%
\authorrunning{E. Olsvik et al.} 
\author{Erlend~Olsvik\inst{1} \and Christian~M.~D.~Trinh\inst{1} \and
Kristian~Muri~Knausgård\inst{2}\textsuperscript{(\Letter)} \and Arne~Wiklund\inst{1} \and Tonje~Knutsen~Sørdalen\inst{3, 4}\and Alf~Ring~Kleiven\inst{3} \and Lei~Jiao\inst{1} \and Morten~Goodwin\inst{1}}
\institute{Centre for Artificial Intelligence Research, 
University of Agder, \\4879, Grimstad, Norway \and Department of Engineering Sciences, University of Agder\and Institute of Marine Research (IMR), His, Norway \and Department of Natural Sciences, Centre for Coastal Research (CCR)\\
University of Agder, Kristiansand, Norway\\
\email{kristianmk@ieee.org}}
\maketitle 
\begin{abstract} 
Our understanding and ability to effectively monitor and manage coastal ecosystems are severely limited by observation methods. Automatic recognition of species in natural environment is a promising tool which would revolutionize video and image analysis for a wide range of applications in marine ecology. However, classifying fish from images captured by underwater cameras is in general very challenging due to noise and illumination variations in water.
Previous classification methods in the literature relies on filtering the images to separate the fish from the background or sharpening the images by removing background noise. This pre-filtering process may negatively impact the classification accuracy. In this work, we propose a Convolutional Neural Network (CNN) using the Squeeze-and-Excitation (SE) architecture for classifying images of fish without pre-filtering. Different from conventional schemes, this scheme is divided into two steps. The first step is to train the fish classifier via a public data set, i.e., Fish4Knowledge, without using image augmentation, named as pre-training. The second step is to train the classifier based on a new data set consisting of species that we are interested in for classification, named as post-training. The weights obtained from pre-training are applied to post-training as a priori. This is also known as transfer learning. Our solution achieves the state-of-the-art accuracy of 99.27\% accuracy on the pre-training. The accuracy on the post-training is 83.68\%. Experiments on the post-training with image augmentation yields an accuracy of 87.74\%, indicating that the solution is viable with a larger data set.
%
%
\end{abstract}
\section{Introduction}\label{Introduction}
Coastal marine ecosystems are highly complex, productive, and important spawning, nursing and feeding areas for numerous of fish species, but studying such biodiversity is often logistically challenging and time-consuming \cite{Perry2018Habitat}\cite{Weinstein2017AComputerVision}. With the recent advancement in cost-effective high definition underwater camera technologies, large volumes of observations from remote areas are now allowing us to test predictions about species’ cryptic behaviour, fundamental ecological processes and environmental changes \cite{PELLETIER201184}. Yet, video data analysis is extremely labour intensive and only a fraction of the available recordings can be analyzed manually, greatly limiting the utility of the data. In addition, accuracy of visual-based assessments is highly dependent on conditions in the underwater environment (depth, light, background noise) and taxonomical expertise in interpreting the videos \cite{Francour1999ComparisonOfFishAbundanceEst}.

Computer vision solutions have been increasingly applied to marine ecology to tackle these problems \cite{li2015fast}\cite{qin2016deepfish}\cite{jin2017deep}. One such solution, the commercial product CatchMeter \cite{white2006automated}, consists of a light box with a camera that photographs and classify the fish as well as provide a length estimate. Fish are recognized by utilizing a threshold for detecting the outline of fish in the images and has a very high classification accuracy of 98.8\%. However, the fish are photographed in a relatively structured environment, which has limited applicability in studies of natural behaviour in the wild.
 
A specific Convolutional Neural Network (CNN) called Fast R-CNN stands out as it applies object detection to extract only the fish from images taken from natural environment, actively ignoring background noise \cite{li2015fast}. The approach starts by pre-training an AlexNet \cite{krizhevsky2012imagenet} on the ImageNet database. The AlexNet is then modified to train on a subset of the Fish4Knowledge data set \cite{fish4knowledge}. As the final step, the Fast R-CNN takes the pre-trained weights and region proposals made by AlexNet as input, and achieves a mean average precision of 81.4\%. In another solution \cite{jin2017deep}, pre-training is applied to a CNN similar to AlexNet. The network consists of five convolutional layers and three fully-connected layers. Pre-training is performed using 1000 images from 1000 categories from the ImageNet data set, and the learned weights are then applied to a CNN after adapting it to the Fish4Knowledge data set. Post-training is then performed using as few as 50 images per category and 10 categories from the Fish4Knowledge data set. The images from the Fish4Knowledge data set are pre-processed using image de-noising. The accuracy achieved on the 1420 test images is 85.08\% using very small amounts of data.

The highest reported accuracy for the Fish4Knowledge data set so far is 98.64\%. The result was achieved by first applying filters to the original images to extract the shape of the fish and remove the background, and then use a CNN with a Support Vector Machine (SVM) classifier function \cite{qin2016deepfish}. The network is named DeepFish, which consists of three standard convolution layers and three fully-connected layers. In addition, previous solutions usually apply a pre-processing of the images in order to remove the noise in the targeted image as much as possible, and to outline the area where fish are located \cite{jin2017deep}\cite{qin2016deepfish}. Although this process can indeed improve the system performance, the set of filters must be chosen carefully, as it may result in a negative performance impact in a live and dynamic scenario. Useful object background information may unintentionally be removed during the segmentation pre-processing, such as indicated by comparison of background discarding Fast R-CNN and background encoding YOLO in \cite{RedmonDGF15}. Considering the noise tolerant nature of CNN with Squeeze-and-Excitation (SE) architecture, it could be an advantage to use the original image to maintain maximum information content.

In this paper, we further explore CNN using the most recent SE architecture, which, to the best of our knowledge, has not previously been utilized in fish classification. In addition to the learning algorithm, we also collect a new data set of temperate fish species in this work. Clearly, the Fish4Knowledge data set is currently limited to tropical fish species. If a CNN is trained on this data set alone, it may not be able to classify fish species in other ecosystems. Therefore, the trained model based on the Fish4Knowledge needs to be further tuned and validated to fit specific ecosystems of interest. Our approach is to first pre-train the network on the Fish4Knowledge data set to learn generic fish features, and then the learned weights from pre-training are adopted as a starting point for further training on the new data set containing images of temperate fish species, which is called post-training. This two-step process is known as transfer learning~\cite{YosinskiTransferLearning}. The solution based on SE-architecture requires no pre-processing of images, except re-sizing to the appropriate CNN input size.

The remainder of the paper is structured as follows. Section~\ref{method} describes the data sets used to train the neural network, and then the detailed network structure and configurations are presented. Section~\ref{experimentalResults} discusses the experimental results for the CNN approach before the work is summarized in the last section.
\section{Data Sets and Deep Learning Approaches}\label{method}
\subsection{The data sets}
Two data sets were used in the test, the Fish4Knowledge data set \cite{li2015fast} and a Norwegian data set with temperate species collected by the research team. Fish4Knowledge is used in pre-training of the neural network, while the temperate data set is used in the post-training. Some differences between the data sets are: (1) The Fish4Knowledge has in addition to the fish images categorized images in trajectories, e.g. a sequence of images taken from the same video sequence or stream. (2) The temperate data set has in addition to the other species a separate folder for male and female Symphodus melops (S. melops). Some individuals of male S. melops have also been tracked and captured by camera multiple times. 
\subsubsection{Fish4Knowledge}
The Fish4Knowledge data set is a collection of images, extracted from underwater videos of fish, off the coast of Taiwan. There is a total of 27230 images cataloged into 23 different species. The top 15 species accounts for 97\% of the images, and the single top species accounts for around 44\% of the images. The number of images for each species range from 25 to 12112 between the species. This creates a very imbalanced data set. Further, the images size ranges from approximately $30\times30$ pixels to approximately $250\times250$ pixels. Another observation in the data set, is that most of the images are taken from a viewpoint along the anteroposterior axis, or slightly tilted from that axis. In that subset of images, most of these images are from the left or right lateral side, exposing the whole dorsoventral body plan in the image. There are some images from the anterior view, but few from the posterior end. Among all the images there were not many images from the true dorsal viewpoint. Most of the selected species have a compressed body plan, e.g. dorsoventral elongate. This creates a very distinct shape when the images are taken from a lateral viewpoint. Hence, images taken from the dorsal view creates a thin, short shape. 
The images also have a background that is relatively light, enhancing the silhouette of the fish.

%
\subsubsection{Temperate fish species}\label{TemperateFishSpecies}
\begin{figure}[ht!]
\begin{center}
\scalebox{0.5}{\includegraphics{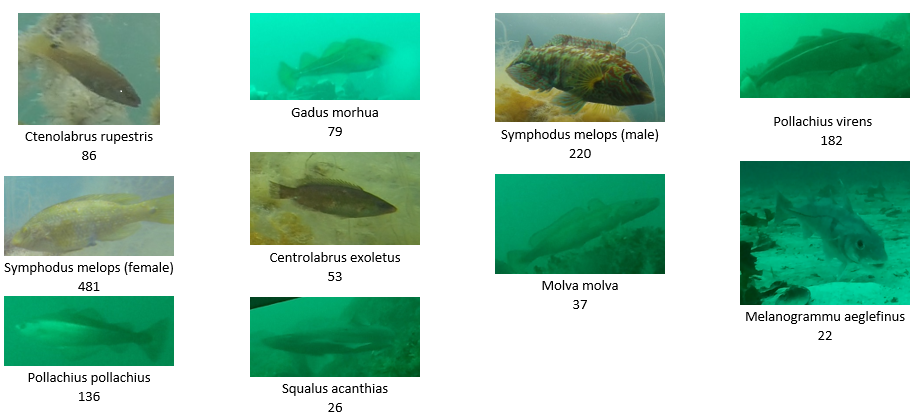}}
\end{center}
\caption{\label{TemperateDatasetFigure} Distribution of the temperate species data set.}
\end{figure}
The temperate data set consists of an image collection of some of the most abundant fish species in Northern Europe. The video recordings were sampled by scientists at the Institute of Marine Research (IMR) in Norway in two different occasions. One part is from video recordings taken from May to June 2015 in a remote and shallow bay on the Austevoll archipelago (Norway, North Sea). GoPro Hero4+ (black) cameras were deployed at 2-5 meters of depth around small reef sites to record the nesting behaviour of \textit{Symphodus melops}. Recording conditions varied between sites and days, especially in sun exposure and background noise. All videos were recorded in full HD resolution of $1920\times1080$ pixels with default settings. Colourful males of \textit{S. melops} build nests to attract females who lay their eggs for the males to care for until they hatch \cite{HalvorsenEtAl2016MaleBiased}. The females are brown in colour and easily distinguished from nesting males. Some males employ a strategy to look indistinguishable from females and do not build nests, but instead sneak’ on other males’ nest \cite{HalvorsenEtAl2017Sex-And-Size-Selective}. Because of the morphological appearances of the different sexes, nest-building males are labelled as ``males" in the data set (accounting for approximately 17\% of the images in the data set), while females and sneaker males are labelled as ``females" (accounting for about 36\%). Two other wrasse species from these videos were also categorized. The second part of the data set was collected with stereo baited remote underwater video (stereo-BRUV). The stereo-BRUV consists of two calibrated GoPro Hero4+ (black) cameras. The cameras were deployed between 8 and 35 meters in 2 coastal areas of Norway: south-estern coast (county of Aust-Agder) and mid-western coast (county of Trøndelag). The stereo-BRUV data sampling is normally used for monitoring marine biodiversity \cite{Mallet2014UnderwaterVideoReview} and  temporal trends in fish assemblages \cite{McLean2011DeclineIn}. A single video frame often contains more than one fish (of same species and/or different species). Differences in depth, visibility, habitat, distance from camera and angle of the fish secured a high variability in pictures of each species. Except from the spiny dogfish (\textit{Squalus acanthias}), the five other species were from the family Gadidae  (\textit{Gadus morhua, Pollachius virens, Pollachius pollachius, Molva molva and Malanogrammus aeglefinus}). Overall, the Norwegian temperate data set has a higher image noise (visibility, background, resolution) and variability of angle of the fish compared with the Fish4Knowledge data set. This is expected to reduce the classification accuracy, but be more realistic for analysis of observations in the natural environment. Fig.~\ref{TemperateDatasetFigure} illustrates a snapshot of the data set.
%
\subsection{CNN-SENet structure}
\begin{figure}[ht]
    \centering
    \scalebox{0.75}{\includegraphics{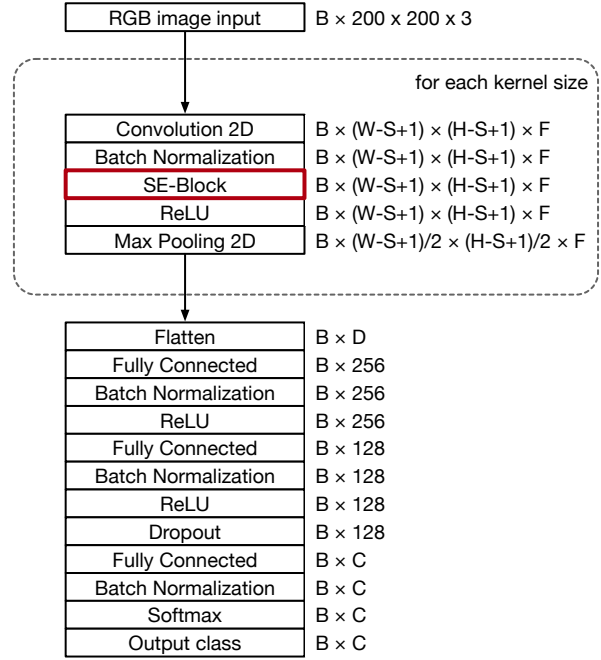}}
    \caption{\label{CNN-SENetFigure}CNN-SENet architecture.}
\end{figure}

\begin{figure}[ht]
    \centering
    \scalebox{0.75}{\includegraphics{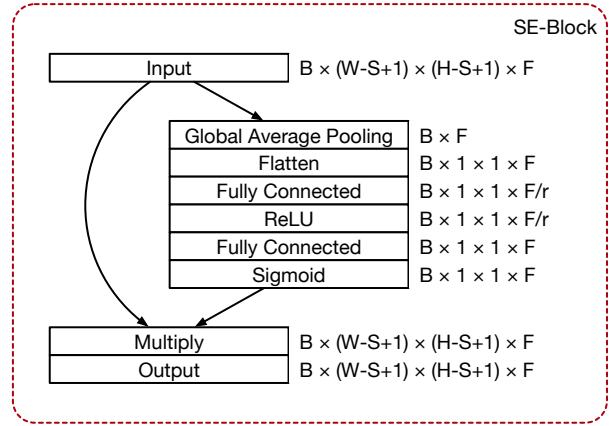}}
    \caption{\label{SE-Block}Squeeze-and-Excitation block.}
\end{figure}

A CNN-SENet, is a Convolutional Neural Network with an added squeeze and excitation (SE) architectural element, that re-calibrates channel wise-feature responses adaptively \cite{HuLiGang2017}.
The architecture of the CNN-SENet, depicted in Fig.~\ref{CNN-SENetFigure}, is configured with the following parameters. Image size in height ($H$), width ($W$) and depth channels; the number of learnable filters ($F$); the batch size ($B$) (default 16), the filter size ($S$), and reduction ratio ($r$) as described in~\cite{HuLiGang2017}. Lastly the number of fish species classifications needs to be added, as parameter $C$.
The input layer takes image of size $200\times200$ with a depth of 3 color channels, R, G, and B. The output is batch normalized before entering the Squeeze-and-Excitation function, called SE block, depicted in Figure~\ref{SE-Block}.
The SE block performs a feature re-calibration through the (1) squeeze operation preventing the network from becoming channel-dependent. This exploits contextual information outside the receptive field and is achieved by doing global average pooling on each input channel before reshaping, and (2) the excitation operation that utilizes the output from the squeeze function by fully capture channel-wise dependencies. This is achieved by the two fully-connected (FC) layers sandwiching the reduction layer, and finally a sigmoid activation layer. Before exiting the SE block, the output from the excitation function is multiplied with the original batch normalized output. This multiplied output is then added to a ReLU layer performing an element-wise activation function, rendering the dimension size unchanged. The output is then sent to a Max Pooling layer, that uses a $2\times2$ filter to reduce and re-size the height and width spatially, rendering an output of $98\times98\times32$. This core portion of the network is stacked to the size of the kernel size, in this case the size of five. The first iteration has a convolutional layer of 32 filters in $5\times5$. The second and third has 64 filters in $3\times3$, the forth 128 filter in $2\times2$,  and the fifth 256 filters in $2\times2$, with all layers applying a horizontal and vertical stride of 1.

Furthermore, the network has 3 FC layers. The first, with 256 neurons, takes the output from the last convolutional layer that is first flattened. The output is then batch normalized before sent to the second FC layer, with 256 neurons. A reduction function is applied after the output from the FC layer is batch normalized. Before entering the last FC layer, with $C$ neurons, a dropout layer of 50\% is applied. The final layer, softmax, applies a classifier function to obtain the probability distribution for each class per input image, using a categorical cross-entropy with the Adam optimizer \cite{kingma2014adam}.

In CNN-SENet, there are certain parameters that need to be configured, including dropout percentage, learning rate, and batch normalization, that are discussed presently. The parameters are configured based on trial-and-error method. For the dropout percentage, clearly, the higher the dropout, the more the information is lost during training because forward- and back-propagation are carried out only on the remaining neurons after dropout is applied. Different percentages of the dropout are tested, and 50\% is configured in this study due to the better overall performance achieved. The learning rates when using the Adam optimizer should be tuned to further optimize the network. After numerous trials, the learning rate is configured as 0.001 without decay. For batch normalization, it has been tested and the results with batch normalization is slightly better than without it. In more details, the accuracy on the testing set without batch normalization is 98.35\%, while the accuracy with batch normalization is 99.27\%. With the above parameters, the model trains faster and has a higher validation accuracy, that concludes the architecture of CNN-SENet. 

To compare CNN-SENet with DeepFish, Table~\ref{SENetVsDeepFishTable} illustrates the main differences between the two. Clearly, CNN-SENet has a more sophisticated structure than DeepFish.
\begin{table}[ht!]
\centering
\caption{Differences between CNN-SENet and DeepFish.}
\begin{tabular}{|l|l|l|}
\hline
 & \textbf{CNN-SENet} & \textbf{DeepFish} \\ \hline
\textbf{Image Size} & $200\times200$ & $47\times47$ \\ \hline
\textbf{Testing Samples} & 4126 & 3098 \\ \hline
\textbf{\begin{tabular}[c]{@{}l@{}}Network Architecture\end{tabular}} & Basic with SE blocks & Basic \\ \hline
\textbf{Classifier} & Softmax & SVM \\ \hline
\textbf{\begin{tabular}[c]{@{}l@{}}Convolutional Layers\end{tabular}} & 5 & 3 \\ \hline
\end{tabular}\label{SENetVsDeepFishTable}
\end{table}
\section{Experiments and results}\label{experimentalResults}
Accuracy and performance of the new fish classification CNN-SENet is quantified and compared with the state-of-the-art networks represented by Inception-V3, ResNet-50 and Inception-ResNet-V2. Additionally, a simplified version of the CNN-SENet, without the Squeeze-and-Excitation blocks, is included to explore how the spatial relationship between fish image colors and other feature layers affect results \cite{HuLiGang2017}.

\subsection{Experiments}
Three different experiments were performed. Pre-training with Fish4Knowledge, post-training with the new temperate Fish Species data set described in subsection~\ref{TemperateFishSpecies} and post-training with an extended version of the new data set using image augmentation techniques. For all three experiments, the applicable data set was divided into 70\% training images, 15\% validation images and 15\% testing images. Both training and validation images are integral parts of the training process, while the testing images were kept out-of-the-loop for independent verification of the ``end product".

All benchmarked networks are trained for 50 epochs with images adapted to their input image size of $200\times200$ RGB pixels, with the notable exception of the $299\times299$ RGB pixels required by Inception-ResNet-V2.
\subsubsection{Pre-training}
\begin{figure}[ht!]
\begin{center}
\scalebox{0.55}{\includegraphics[trim=0cm 0.8cm -2.2cm 1.0cm]{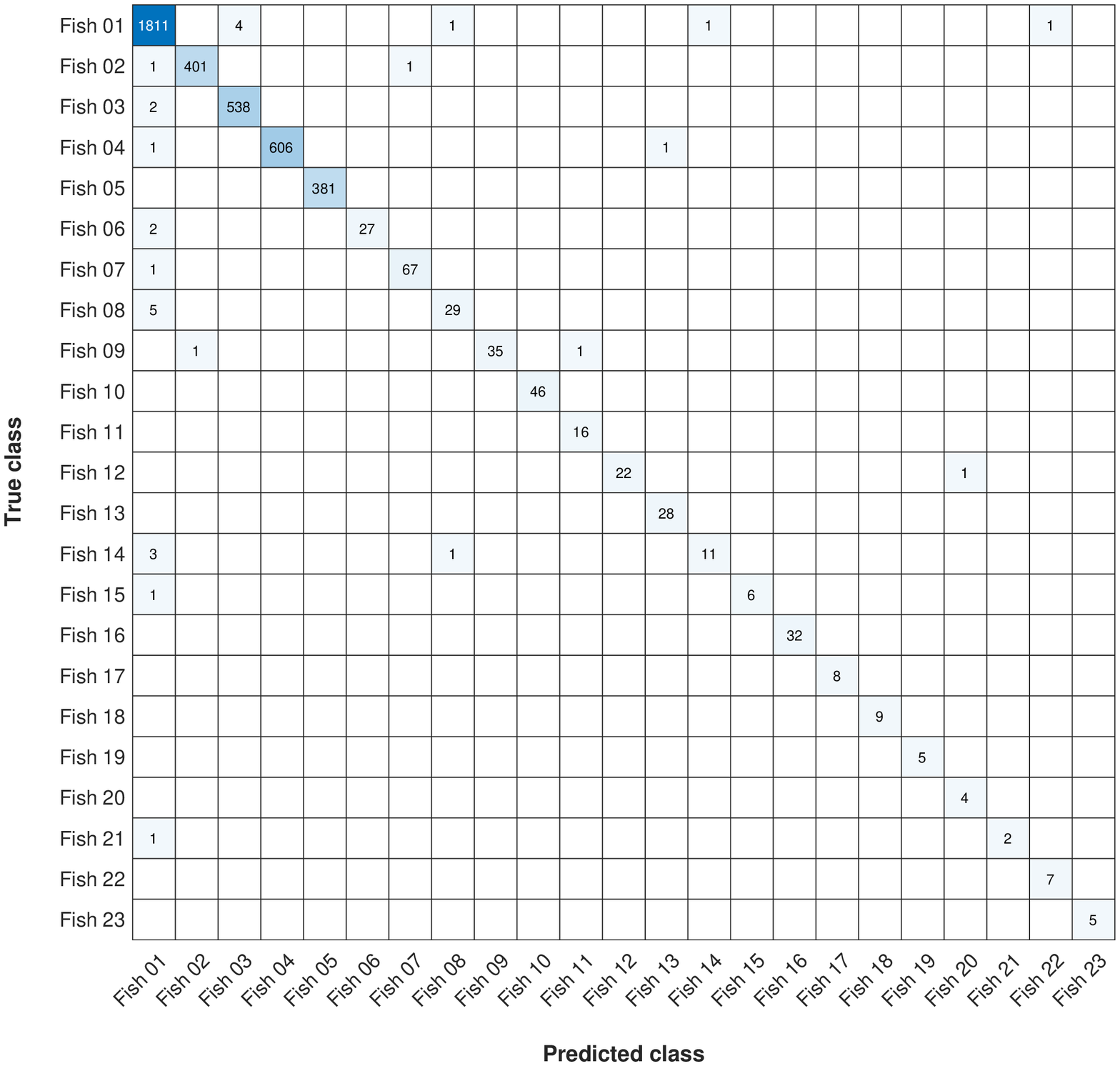}}
\end{center}
\caption{\label{ConfusionPreFigure} Confusion matrix for Fish4Knowledge data set pre-training with CNN-SENet.}
\end{figure}
Pre-training was performed using a data set consisting of 19149 Fish4Knowledge images, with an additional 4126 images for verification and 4126 images reserved for testing. The selected training configuration consists of a single run with 50 training epochs and a batch size of 16. Results from pre-training are evaluated using weights from the epoch with highest validation accuracy, and not necessarily the final epoch.
\begin{table}[ht!]
\centering
\caption{Testing accuracy and time per epoch on pre-training.}
\begin{tabular}{|c|c|c|}
\hline
\textbf{Network}                     & \textbf{Testing Accuracy}		& \textbf{Time One Epoch} \\ \hline
\textbf{Inception-V3}   &    99.18\%    &923 s                \\ \hline
\textbf{ResNet-50}      &   98.86\%     &646 s                 \\ \hline
\textbf{Inception-ResNet-V2}    &    98.59\% & 2221 s                  \\ \hline
\textbf{CNN-SENet}      &    99.27\%  & 197 s                 \\ \hline
\textbf{\begin{tabular}[c]{@{}l@{}}CNN-SENet without\\ Squeeze-and-Excitation\end{tabular}}      &   99.15\%    & 159 s                    \\ \hline
\end{tabular}\label{PreTrainingAccuracyTable}
\end{table}
\subsubsection{Post-training}
\begin{figure}[ht]
\begin{center}
\scalebox{0.45}{\includegraphics[trim=0cm 0.8cm -2.1cm 0cm]{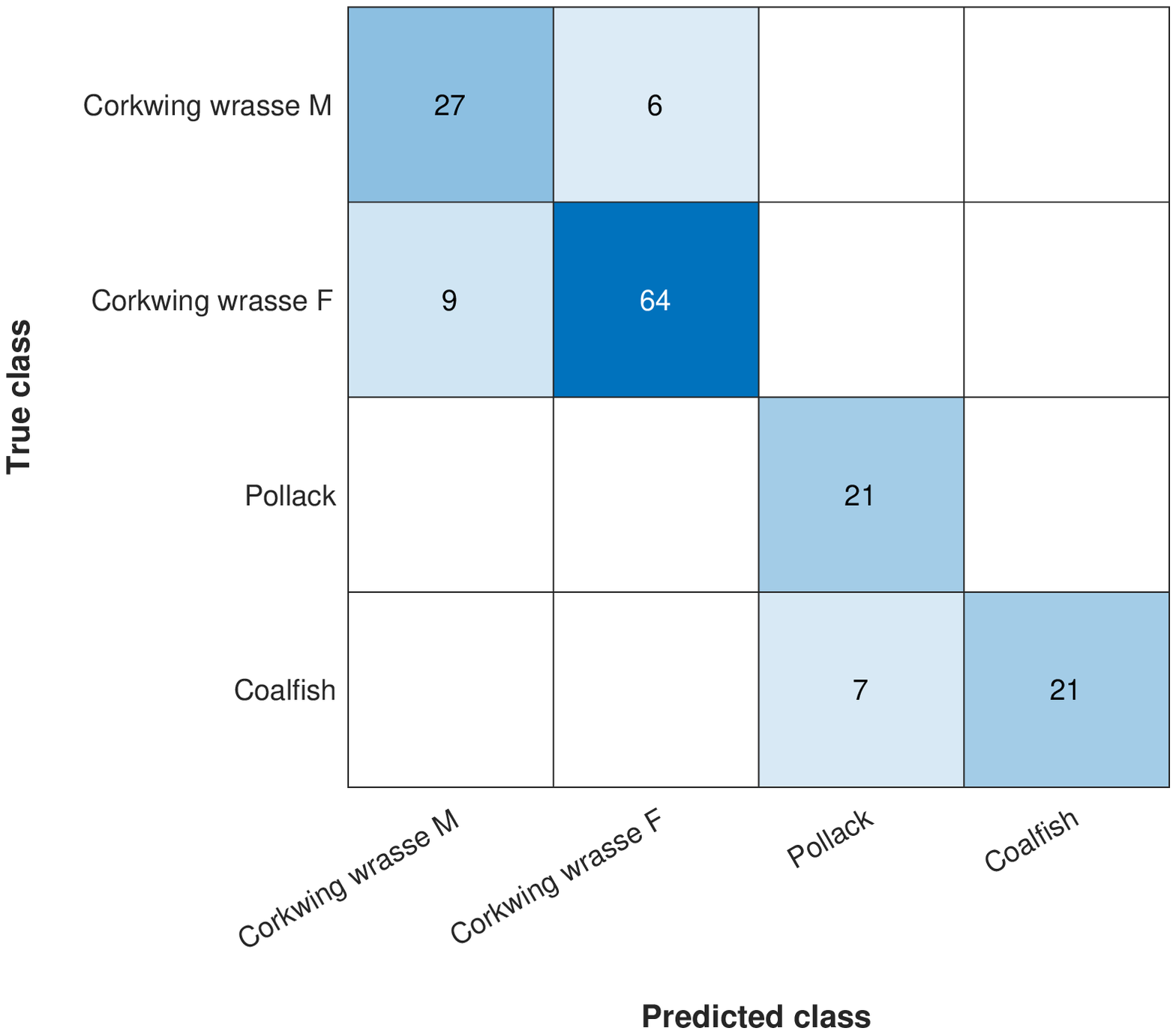}}
\end{center}
\caption{\label{ConfusionPostFigure} Confusion matrix for temperate data set post-training with CNN-SENet.}
\end{figure}
Post-training was performed using 712 images of four fish classes from the temperate fish species data set described in section~\ref{TemperateFishSpecies}. An additional 155 images was used for verification during training, and a subset of 155 images of the same classes were reserved for testing. Corkwing wrasse (male), Corkwing wrasse (female), Pollach and Coalfish were selected for the experiment as a reasonable number of images of different individuals under varying conditions was available for these species.

The post-training process consists of 50 epochs and a batch size of 8. The batch size was reduced, compared to pre-training, to compensate for the relatively small number of available temperate fish images. Weights from the pre-training step are loaded before initiating post-training, and post-training accuracy is evaluated using the weights from the final epoch.

The rationale for this post-training method is to make use of the more or less generic fish identification features learned from the large Fish4Knowledge data set. Post-training will then start with the network in a ``fish-class-sensitive" state and proceed by learning specific features of the temperate species on top of this.

Fish4Knowledge consists of images of 23 different classes. The selected subset of the temperate data set consists of 4 classes. To prepare the loaded pre-trained model for post-training, the last fully connected (FC) layer with 23 output neurons, suitable for 23 fish classes, is replaced with a similar layer with 4 output neurons.
\begin{table}[ht]
\centering
\caption{Average testing accuracy over 10 runs and time per epoch on post-training.}
\begin{tabular}{|c|c|c|}
\hline
\textbf{Network}                     & \textbf{Testing Accuracy}		& \textbf{Time One Epoch} \\ \hline
\textbf{Inception-V3} &    85.42\%         & 33 s              \\ \hline
\textbf{ResNet-50}    &    82.39\%        & 47 s               \\ \hline
\textbf{Inception-ResNet-V2} & 78.84\%  	& 91 s                \\ \hline
\textbf{CNN-SENet}       &    83.68\%       & 9 s           \\ \hline
\textbf{\begin{tabular}[c]{@{}l@{}}CNN-SENet without\\ Squeeze-and-Excitation\end{tabular}}       &    82.32\%       & 7 s           \\ \hline
\end{tabular}\label{PostTrainingAccuracyTable}\end{table}
\subsubsection{Post-training with Image Augmentation}\label{ImageAugmentationExperiment}
Data augmentation techniques in machine learning aims at reducing overfitting problems by expanding a data set (base set) by introducing label-preserving transformations. For an image data set, this means that transformed copies of the original images in the base set are produced. These additional training data enables a network under training to learn more generic features, by reducing sensitivity to augmentation operations that transforms the image but not severely the characterizing visual features of for example a fish. \cite{KrizhevskyImagenet2012ImageAugmentation}

The main algorithm flow is the same as for the post-training version, but the data set was expanded by using the following transformation operations. Images are rotated randomly within a specific range, according to an uniform distribution. Images are vertically and horizontally shifted a random fraction of the image size. Scaling and shearing transformations are applied randomly, and lastly half of the images are flipped horizontally.

\subsection{Results}

\subsubsection{Pre-training}
Results from pre-training on Fish4Knowledge are presented in Table~\ref{PreTrainingAccuracyTable}. The testing accuracy is on par with or exceeds the level of accuracy achieved with previous state-of-art solutions described in section~\ref{Introduction}.

CNN-SENet with Squeeze-and-Excitation achieves 99.15\% test accuracy, almost identical results as the Inception-V3 algorithm when it comes to accuracy. However, the run time for each epoch is roughly three times larger for Inception-V3. The training-runtime is expected to be reflected in prediction. CNN-SENet without Squeeze-and-Excitation is faster than the SE-version, but also slightly less accurate during these tests.

Inception-ResNet-V2 achieves the lowest test accuracy and also the highest time consumed for each epoch during training. The required input image size is $299\times299$, compared to $200\times200$ for the other networks under test. As the required resolution is higher than the resolution of most Fish4Knowledge images, the necessary upscaling process may negatively affect accuracy. Additionally, the larger input size also dramatically increases the computational complexity and leads to longer time on each epoch.

A confusion matrix for the CNN-SENet pre-training run is included as shown in Fig.~\ref{ConfusionPreFigure}. Fish 01 seems to attract more wrong predictions than the other species. The reason for this is unknown, but the imbalance in the data set could explain some of the behavior, as the ability to learn Fish 01 will be more rewarding during training as it occurs more frequently.

\subsubsection{Post-training with and without image augmentation}
Results from the post-training experiment indicates that this is a more challenging image recognition task. Without image augmentation, the highest average testing accuracy achieved was 85.42\% using the Inception-V3 CNN algorithm as listed in Table~\ref{PostTrainingAccuracyTable}. CNN-SENet performance is few percent below, but with a significantly better training time for each epoch. All bench-marked algorithms show significantly reduced accuracy compared to the results from pre-training. The temperate species data set used for post-training is challenging, in the sense that it contains few images overall. The data set also consists of pictures of fish under low visibility conditions, and situations where the fish silhouette is not always prominent.

Image augmentation, as described in section~\ref{ImageAugmentationExperiment}, improves the results for post-training for all benchmarked algorithms, as shown in Table~\ref{PostTrainingWithImageAugmentationAccuracyTable}. The ResNet-50 network reaches just above 90\% testing accuracy. CNN-SENet accuracy increases approximatly four percentage points compared to post-training without image augmentation. The training time for each epoch does not change notably using image augmentation, so the metric was omitted from Table~\ref{PostTrainingWithImageAugmentationAccuracyTable}.
\begin{table}[ht]
\centering
\caption{Average testing accuracy over 10 runs on post-training with image augmentation.}
\begin{tabular}{|c|c|}
\hline
\textbf{Network}                     & \textbf{Testing Accuracy}	\\ \hline
\textbf{Inception-V3} &    88.45\%      \\ \hline
\textbf{ResNet-50}    &    90.20\%         \\ \hline
\textbf{Inception-ResNet-V2} & 82.39\%       \\ \hline
\textbf{CNN-SENet}       &    87.74\%   \\ \hline
\textbf{\begin{tabular}[c]{@{}l@{}}CNN-SENet without\\ Squeeze-and-Excitation\end{tabular}}       &    83.55\%       \\ \hline
\end{tabular}\label{PostTrainingWithImageAugmentationAccuracyTable}
\end{table}
\section{Conclusions}\label{conclusion}
We propose a Convolutional Neural Network implementing the Squeeze-and-Excitation (CNN-SE) architecture, which is specifically tuned and trained for biometric classification of fish. The experimental results show that CNN-SENet achieves the state-of-the-art accuracy of 99.27\% on the Fish4Knowledge data set without any data augmentation or image pre-processing. For post-training, where the CNN-SENet is specialized for recognizing temperate fish species, the achieved average accuracy is 83.68\%. The lower accuracy can be explained by the small size of the new temperate species data set combined with high variation in image data.  For both approaches, CNN-SENet with SE blocks has a higher accuracy than without the SE blocks, indicating that SE has a positive effect on accuracy. In conclusion, we show that CNN with SE architecture is a powerful and effective tool for automatic analysis of fish images taken in the the wild, but future work should make use of much larger and well-labelled data sets.

\bibliographystyle{splncs04}

\end{document}